\documentclass[10pt,twocolumn,letterpaper]{article}

\usepackage[pagenumbers]{cvpr} 

\usepackage{graphicx}
\usepackage{amsmath}
\usepackage{amssymb}
\usepackage{booktabs}

\usepackage[pagebackref,breaklinks,colorlinks]{hyperref}

\usepackage[capitalize]{cleveref}
\crefname{section}{Sec.}{Secs.}
\Crefname{section}{Section}{Sections}
\Crefname{table}{Table}{Tables}
\crefname{table}{Tab.}{Tabs.}

\def\cvprPaperID{*****} 
\def\confName{CVPR}
\def\confYear{2022}

\begin{document}

\title{Improving Object Detection in Medical Image Analysis through Multiple Expert Annotators: An Empirical Investigation}

\author{Hieu H. Pham \\ Coordinated Science Laboratory, UIUC \\ VinUni-Illinois Smart Health Center \& CECS, VinUniversity  \\ \texttt{hieu.ph@vinuni.edu.vn}
\and
Khiem H. Le \\ VinUniversity \\ \texttt{khiem.lh@vinuni.edu.vn}
 \and
 Tuan V. Tran \\  VinUniversity \\ \texttt{tuan.tv@vinuni.edu.vn}
 \and
Ha Q. Nguyen \\ Vingroup Big Data Institute \\ \texttt{v.hanq3@vinbigdata.org}
}

\maketitle
\section{Introduction}  

The work discusses the use of machine learning algorithms for anomaly detection in medical image analysis and how the performance of these algorithms depends on the number of annotators and the quality of labels. To address the issue of subjectivity in labeling with a single annotator, we introduce a simple and effective approach that aggregates annotations from multiple annotators with varying levels of expertise. We then aim to improve the efficiency of predictive models in abnormal detection tasks by estimating hidden labels from multiple annotations and using a re-weighted loss function to improve detection performance. Our method is evaluated on a real-world medical imaging dataset and outperforms relevant baselines that do not consider disagreements among annotators. This work\footnote{This is a short version submitted to the Midwest Machine Learning Symposium (MMLS 2023), Chicago, IL, USA.}  has been accepted for publication by IEEE Access  and its full form can be found in \cite{le2023learning}. We also released the dataset and the code used in this study\footnote{\url{https://vindr.ai/datasets/cxr}.}.

\section{Learning from multiple annotators} 

\begin{figure}[!ht]
  \centering
  \includegraphics[height=70pt]{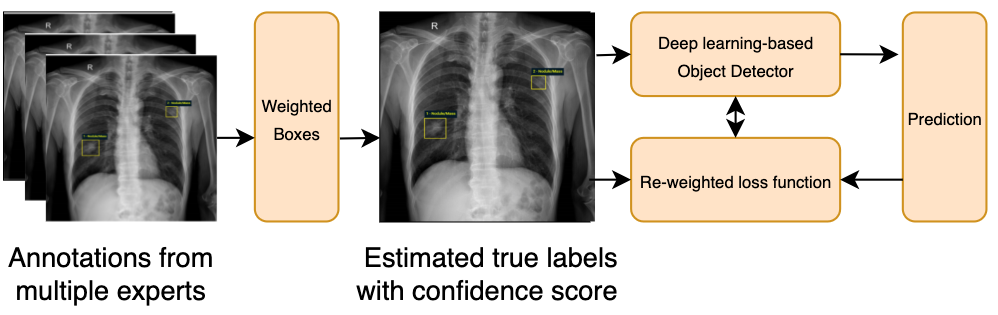}
  \caption{Illustration of the proposed approach.}
  \label{fig:main_diagram}
  
\end{figure}

Existing works that are highly related to our work, including learning from multiple annotators \cite{10.14778/3055540.3055547, Sheng_Zhang_2019, jin2020technical, tanno2019learning, li2021learning, HumanError2020,10.1145/1401890.1401965, NIPS2011_c667d53a, Dawid1979MaximumLE, 10.5555/2998687.2998822, whitehill2009whose, 10.1145/1553374.1553488, ramakrishna2016expectation, rodrigues2018deep, NEURIPS2020_b5d17ed2, tanno2019learning } and weighted training techniques \cite{1467360,shrivastava2016training,schapire2013explaining,lin2017focal,10.1145/1553374.1553380,zhang2021understanding,10.5555/3305381.3305406}. Unlike any approaches above, we aggregated annotations from multiple annotators and propose a re-weighted loss function that assigns more weights to more confident examples that determine by the consensus of multiple annotators. Figure~\ref{fig:main_diagram} shows an overview of our method. Specifically, we first estimate the actual labels using WBF algorithm~\cite{Solovyev_2021}. They are then used to train a typical object detector with a re-weighted loss function. Note a general form of the loss function for those detectors can be written as

\begin{gather}
\label{eq:general_loss}
\begin{aligned}
\mathcal{L}\left(p, p^{*}, t, t^{*}\right) &=\mathcal{L}_{cls}\left(p, p^{*}\right)+\beta I(t) \mathcal{L}_{loc}\left(t, t^{*}\right) \\
I(t) &=\left\{\begin{array}{ll}
1 & \text {if} \operatorname{IoU}\left\{a, a^{*}\right\}>\eta \\
0 & \text { otherwise}.
\end{array}\right.
\end{aligned}
\end{gather}

where $t$ and $t^*$ are the predicted and ground truth box coordinates, $p$ and $p^*$ are the class category probabilities, respectively; $\operatorname{IoU}\left\{a, a^{*}\right\}$ denotes the Intersection over Union (IoU) between the anchor $a$ and its ground truth $a^{*}$; $\eta$ is an IoU threshold for objectness, i.e. the confidence score of whether there is an object or not; $\beta$ is a constant for balancing two loss terms $\mathcal{L}_{cls}$ and $\mathcal{L}_{loc}$ \cite{zou2019object}. We use fused boxes confidence scores $c^i_k$ obtained from WBF algorithm~\cite{Solovyev_2021} to get a re-weighted loss function that emphasizes boxes with high annotators agreement. The new loss function can now be written as
\begin{equation}
\label{eq:earl}
\mathcal{L}\left(p, p^{*}, t, t^{*}\right) = c \mathcal{L}_{cls}\left(p, p^{*}\right) + c \beta I(t) \mathcal{L}_{loc}\left(t, t^{*}\right).
\end{equation}

\section{Experiments and Results} 
We validated our method on VinDr-CXR~\cite{nguyen2020vindr}, a real-world chest X-ray dataset with labels provided by multiple radiologists for a typical task medical imaging \cite{nguyen2021vindr,tran2021learning,pham2021dicomx,nguyen2021clinical,pham2021interpreting,nguyen2022vindrd,nguyen2022vindrmamo,nguyen2022deployment,pham2022accurate,nguyen2022learning}. It consists of 18,000 chest X-ray scans, with 15,000 for training and 3,000 for testing sets. We compared the performance of the proposed method against (1) \texttt{Baseline} using all annotations as the ground truth; models trained on annotations provided by each annotator independently (\texttt{Annotators \#1, \#2, \#3}) and (3) an ensemble of independent models trained on separate radiologists' annotation sets. Table \ref{tab:vindr_cxr} reports the experimental results in mAP@0.4 score. Our method outperforms the baselines, individual models, and even the ensemble of individual experts' models. Experimental results validated the effectiveness of the proposed method.

\begin{table}[!ht]
\centering
\setlength{\tabcolsep}{11.2pt}
\renewcommand{\arraystretch}{1.1}
\caption{\label{tab:vindr_cxr} Experimental results on the VinDr-CXR dataset.\\[-0.3cm]}
\small{
 \begin{tabular}{lc}
\hline
                      \textbf{Method} & \textbf{mAP@0.4} \\ \hline
 \texttt{Baseline} & 0.148            \\
                                \texttt{Annotator \#1}             & 0.121            \\
                                \texttt{Annotator \#2}            & 0.132            \\
                                \texttt{Annotator \#3}           & 0.124            \\
                                \texttt{Ensemble of all annotators}     & 0.154            \\
                                \texttt{\textbf{Ours}}   & \textcolor{red}{0.158} \\ \hline

\end{tabular}
}
\end{table}
\section{Discussions and Conclusion}
The proposed method is the first effort to train an image detector from labels provided by multiple annotators. We empirically showed a notable improvement in terms of mAP scores by estimating the true labels and then integrating the implicit annotators' agreement into the loss function to emphasize the accurate bounding boxes over the imprecise ones. The idea is simple but effective, allowing the overall framework can be applied in training any machine learning-based detectors. However, the overall architecture is not end-to-end. It may not fully exploit the benefits of combining truth inference and training the desired image detector.

\bibliographystyle{ieee_fullname}
\bibliography{references}

\end{document}